# On the mathematic modeling of non-parametric curves based on cubic Bézier curves

*Ha Jong Won, Choe Chun Hwa, Li Kum Song*

College of Computer Science, **Kim Il Sung** University, Pyongyang, DPR of Korea

**Abstract** Bézier splines are widely available in various systems with the curves and surface designs. In general, the Bézier spline can be specified with the Bézier curve segments and a Bézier curve segment can be fitted to any number of control points. The number of control points determines the degree of the Bézier polynomial.
This paper presents a method which determines control points for Bézier curves approximating segments of obtained image outline(non-parametric curve) by using the properties of cubic Bézier curves. Proposed method is a technique to determine the control points that has generality and reduces the error of the Bézier curve approximation. Main advantage of proposed method is that it has higher accuracy and compression rate than previous methods. The cubic Bézier spline is obtained from cubic Bézier curve segments. To demonstrate the various performances of the proposed algorithm, experimental results are compared.

### 1. Introduction

In the practices for computer vision application recently, objects are captured using scanner and camera. Often the objects are represented by their boundaries. But in image objects are represented by their outlines. The outlines are non-parametric curves and so for storage in computer memory as an image file there are some challenges determining the mathematic models approximating these outlines. Outline shapes of natural objects are approximated with the various spline models such as Hermite curves, Bézier curves and B-spline curves [1]. There are techniques using various spline models like Bézier splines [2,3], B-splines [4], Hermite interpolations [5]. [6,7,8] proposed methods to determine the control points for the cubic Bézier curve approximation to the given curve. The outline is divided into several segments and cubic Bézier curve approximation is then performed over each segment [6,7]. Cubic Bézier spline that is obtained does not involve in least-squares fit or error minimization. The control points for approximating Bézier curve are determined by a search algorithm [6,7,8]. The subdivision is done by recursive algorithm during the approximation process for decomposition of outline into smaller curves. [6,7,8] proposed methods to determine the control points for the Bézier curve approximation to the given curve, but the error of curve approximation is large because of generality. The various areas of multimedia technology involve in shape description of characters [9] and objects [10,11,12], active shape lip modeling [13], shape coding and error concealment for MPEG-4 coded objects [14] and surface mapping [15,11]. [14,16] present a dynamic Bézier curve model by a parametric shift of the Bézier curve points in the gap between the curve and its control polygon. The value of shifting parameter is dynamically determined and dynamic Bézier curve retains the properties of the Bézier curve. The other techniques of curve approximation are using control parameters [5], genetic algorithms [17], and wavelets [18]. In above techniques, the error of the curve approximation was reduced or computation time was reduced. But still, the most of the techniques were based on calculating and minimizing approximation error.

In this paper we proposed a technique to determine the control points that has generality and reduces the error of the Bézier curve approximation.

This paper is organized as follows. Section II explains the cubic Bézier curves and its properties. Section III describes the segmentation of given outline by using corner detection. Section IV derives the process of control point estimation for cubic Bézier curve approximation. Finally, we report our experimental results in section V and conclude in section VI.

## 2. Cubic Bézier curve and properties

In general, the Bézier curve can be fitted by any number of control points.

The number of control points to be approximated and their relative position determine the degree of the Bézier curve polynomial. Bézier curve can be specified with blending functions. Cubic Bézier curve $P(u)$ is generated with four control points $P_i = (x_i, y_i), i = \overline{0,3}$ and four blending functions $B_i(u), i = \overline{0,3}$ [1].

This Bézier polynomial function is represented by following equation:

$$P(u) = P_0 B_0(u) + P_1 B_1(u) + P_2 B_2(u) + P_3 B_3(u), \quad 0 \leq u \leq 1 \tag{1}$$

where the blending functions $(B_0, B_1, B_2, B_3)$ are the four Bernstein polynomials for cubic Bézier curve:

$$\left.\begin{array}{l} B_0(u) = (1-u)^3 \\ B_1(u) = 3u(1-u)^2 \\ B_2(u) = 3u^2(1-u) \\ B_3(u) = u^3 \end{array}\right\} \tag{2}$$

Plots of the four cubic blending functions are given in Fig. 1.

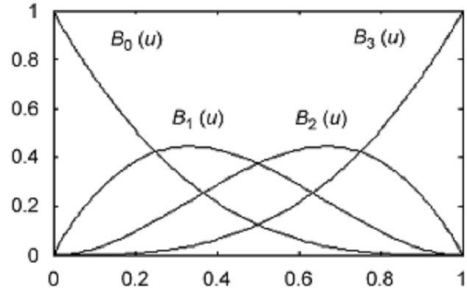

Fig. 1. Blending functions of cubic Bézier curve

The shape of the blending functions determine how the control points influence the shape of the curve for values of parameter $u$ over the range from 0 to 1.

Properties of cubic Bézier curve are represented as following:

• Cubic Bézier curve always passes through the first and last control points.

• The slope at the beginning point of the cubic Bézier curve is along the line $P_0 P_1$ joining the first two control points, and the slope at the end of the curve is along the line $P_2 P_3$ joining the last two endpoints.

• Cubic Bézier curve always lies within the convex hull of the control points. The blending functions of cubic Bézier curve are all positive and their sum is always 1:

$$\sum_{i=0}^{3} B_i(u) = 1 \tag{3}$$

• The blending function of cubic Bézier curve $B_0$ is nonzero function with a value of 1 at $u = 0$ and $B_3$ is nonzero function with a value of 1 at $u = 1$, they are symmetrical relatively to $u = 0.5$.

• Blending function of cubic Bézier curve $B_1$ is maximum at $u = 1/3$, and $B_2$ is maximum at $u = 2/3$ and they are symmetrical relatively to $u = 0.5$.

• Blending functions $B_1$ and $B_2$ influence the shape of the curve at intermediate values of parameter $u$ over

the range from 0 to 1.

## 3. Outline segmentation

Arbitrary points on cubic Bézier curve except two end points ensure $C^2$ continuity. And some points on outline must ensure only $G^0$ continuity, not $C^1$ or $C^2$. Therefore these points must break points between two cubic Bézier curves. This is most important reason of outline segmentation. Next, because subdividing of curve uses a lot of computation in calculating and minimizing approximation error, segmentation of outline is also needed for simplifying the fitting process. This is second reason of outline segmentation. Break points on boundary can be obtained using a suitable corner detector. Various corner detection algorithms are proposed for outline modelng[19,20,21,22,6]. We use the method proposed in [6] for comparative study. This algorithm is briefly described below (readers are referred to [20] for details). Firstly, for all contour point $P_i$, $1 \leq i \leq n$, where $n$ is the number of points in a closed loop curve, $P_k$ is calculated as

$$P_k = P_{(i+L) \bmod n}, \qquad (4)$$

where $L$ is a length parameter which represents the length of region of support. Region of support is addressed by various works [21,23,24]. Setting its length too short tends to be irregular curve and increasing it tends to smooth a sharp corner. In our study, $L$ is taken as 14 pixels as in [6].

Secondly, perpendicular distances of all contour points between $P_i$ and $P_k$ are calculated from the straight line joining these two points. If point $P_j$ has maximum perpendicular distance and its perpendicular distance $d_j$ is greater than threshold $D$, $P_j$ is selected as a candidate corner point and the distance $d_j$ is assigned to $P_j$. If there are more than one point $P_j$ with maximum perpendicular distance, all points with maximum perpendicular distance are selected as candidate corner points. Distance parameter $D$ prevents selection of the locally sharp point and value of $D$ is taken as a length equal to 2.6 pixels as in [6]. The perpendicular distance $d_j$ from point $P_j$ to the straight line joining the point $P_i$ and $P_k$ can be calculated as following:

$$\begin{aligned}
&\text{If } m_x = 0 \\
&\text{Then } d_j = |P_{j,x} - P_{i,x}| \\
&\text{Else } d_j = \frac{|P_{j,y} - mP_{j,x} + mP_{i,x} - P_{i,y}|}{\sqrt{m^2 + 1}} \\
&\text{where } m_x = P_{k,x} - P_{i,x}, \; m = \frac{m_y}{m_x} = \frac{P_{k,y} - P_{i,y}}{P_{k,x} - P_{i,x}}
\end{aligned} \qquad (5)$$

Thirdly, all candidate corner points are detected for $i = 1$ to $n$. If several distances are assigned to a corner point $P_j$, the highest is selected as distance $d_j$ of $P_j$. Fourthly, the candidate corner is thrown away if any other candidate with higher value of $d_j$ is in the range $R$. That is, the candidate with highest value of $d_j$ among $R$ number of points on its both sides is the corner point. If there is more than one candidate corner point with highest value of $d_j$, one of them is selected as corner point. Default value of $R$ is equal to $L$ like [6]. Outline is divided into different curves by obtained corners and cubic Bézier curve is determined for each segment.

**4. Estimation of control points about Bézier curve segment that approximates outline**

Control point estimation of cubic Bezier curve is to determine suitable positions of four control points from a given curve.

Let's determine the control points for cubic Bézier curve approximating of the outline segments, when outlines in input image are determined and $P_l$, $l = \overline{1, n}$ are the corner points and $C_{l-1,i}$, $i = \overline{1, m}$ are the contour points which are detected in the outline.

We determine the control points of cubic Bézier curve that approximates to the outline segment with two corner points as follows.

If there are outline segments, which two end points are corner points($P_l$, $l = \overline{1, n}$) detected on the outline $C$, then control points of the cubic Bézier curve on the $l$ th outline segment with the contour points are as follows:

$$P_{l_0} = P_l, \quad P_{l_3} = P_{l+1}, \quad (6)$$

$$P_{l_1} = \frac{c_1(t')B_{l_1}(t') - c_2(t')B_{l_2}(t')}{B^2_{l_1}(t') - B^2_{l_2}(t')}, \quad P_{l_2} = \frac{c_2(t')B_{l_1}(t') - c_1(t')B_{l_2}(t')}{B^2_{l_1}(t') - B^2_{l_2}(t')}, \quad (7)$$

where $P_{l_i}$, $i = \overline{0,3}$ is $i$ th control point of $l$ th outline segment,

$$\begin{aligned} c_1(t') &= C(t') - P_{l_0}B_{l_0}(t') - P_{l_3}B_{l_3}(t'), \\ c_2(t') &= C(1-t') - P_{l_0}B_{l_0}(1-t') - P_{l_3}B_{l_3}(1-t') \end{aligned} \quad (8)$$

and $B_{l_0}(t')$, $B_{l_1}(t')$, $B_{l_2}(t')$, $B_{l_3}(t')$ are cubic Bézier blending functions.

$C(\cdot)$ is contour point.

We can consider the reason as follows.

Let the corner points detected on the outline be $P_l$, $l = \overline{1, n}$, the contour points between the corner points be $C_{l-1,i}$, $i = \overline{1, m}$, $C_{l-1,1} = P_l$, $C_{l-1,m} = P_{l+1}$ and let's use the following notation:

$$t' = L_{l_i} / L_l, \quad (9)$$

where $L_l = |P_{l+1} - P_l|$, $L_{l_i} = |C'_{l-1,i} - P_l|$, $i = \overline{1, m}$, $C'_{l-1,i}$ is projection of contour point to line joining two end points of outline segment.

Generally, the cubic Bézier curve with respect to the variable $t'$ is written by following equation:

$$C(t') = P_{l_0}B_{l_0}(t') + P_{l_1}B_{l_1}(t') + P_{l_2}B_{l_2}(t') + P_{l_3}B_{l_3}(t'), \quad 0 \leq t' \leq 1. \quad (10)$$

For the variable $1-t'$, the cubic Bézier curve is written as follows:

$$C(1-t') = P_{l_0} B_{l_0}(1-t') + P_{l_1} B_{l_1}(1-t') + P_{l_2} B_{l_2}(1-t') + P_{l_3} B_{l_3}(1-t'). \qquad (11)$$

Let's use the following symbols:

$$\begin{aligned} c_1(t') &= C(t') - P_{l_0} B_{l_0}(t') - P_{l_3} B_{l_3}(t'), \\ c_2(t') &= C(1-t') - P_{l_0} B_{l_0}(1-t') - P_{l_3} B_{l_3}(1-t') \end{aligned} \qquad (12)$$

Then

$$\begin{aligned} P_{l_1} B_{l_1}(t') + P_{l_2} B_{l_2}(t') &= c_1(t'), \\ P_{l_1} B_{l_1}(1-t') + P_{l_2} B_{l_2}(1-t') &= c_2(t') \end{aligned} \qquad (13)$$

And from properties of cubic Bézier blending function the following relations can be obtained:

$$B_{l_1}(t') = B_{l_2}(1-t'), \quad B_{l_1}(1-t') = B_{l_2}(t'). \qquad (14)$$

Hence, we can obtain

$$\begin{aligned} P_{l_1} B_{l_1}(t') + P_{l_2} B_{l_2}(t') &= c_1(t'), \\ P_{l_1} B_{l_2}(t') + P_{l_2} B_{l_1}(t') &= c_2(t') \end{aligned} \qquad (15)$$

Therefore, the control points $P_{l_1}$ and $P_{l_2}$ are calculated by

$$P_{l_1} = \frac{c_1(t') B_{l_1}(t') - c_2(t') B_{l_2}(t')}{B^2_{l_1}(t') - B^2_{l_2}(t')}, \quad P_{l_2} = \frac{c_2(t') B_{l_1}(t') - c_1(t') B_{l_2}(t')}{B^2_{l_1}(t') - B^2_{l_2}(t')}. \qquad (16)$$

From above result, cubic Bézier curve for outline segment is determined by equation (10) using two end points of outline segment and two middle control points obtained by equation (7).

## 5. Result analysis and comparative study

Main advantage of proposed algorithm is higher accuracy and compression rate than previous methods. To verify this advantage we do a comparison experiment. We use experiment data that was used in [6,23,24] for objectivity of comparison. Firstly, we examine control point spread of cubic Bézier curve for a segment of closed loop curve.

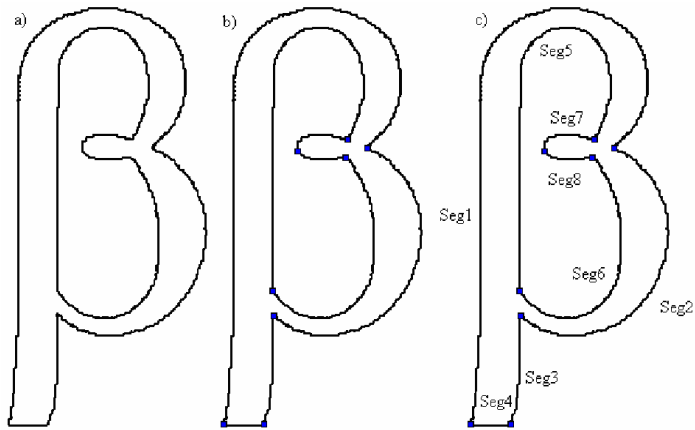

Fig. 2. Outline segmentation: a) original outline, b) detected corners (blue) and c) assignment of segment numbers.

Fig. 2 shows outline of character "$\beta$", detected corners and numbers of each segments. Outline of Fig. 2a was extracted from image of 180×372 resolution. Blue points in Fig. 2b are 8 corners that were extracted by corner detection algorithm. In the experiment, $L$ and $R$ are taken as 14, $D$ is taken as 2.6 as in [6]. Fig. 2c shows numbers of each segment. For comparison experiment, we apply the algorithm [6] and the proposed algorithm to seg2. First we do Bézier curve approximation of curve segment without subdivision of curve. Two control points of Bézier curve are decided as means of obtained control point spreads. Fig. 3 shows curve generated by whole control point spread and control point spread decreased variance. (a), (b), (c) show the result of method [6] and (d), (e), (f) result of proposed method. In Fig. 3, black curve is image outline, red curve is approximated Bézier curve, blue points are two end points of curve, black points are candidate control point spread and yellow points are Bézier control points. First column is whole control point spread, its mean and Bézier curve generated by it. Second column is control point spread that 5% candidate with maximum variance is removed once, its mean and Bézier curve generated by it and third column is control point spread that 5% candidate with maximum variance is removed twice, its mean and Bézier curve generated by it. Table 1 shows numerical result of experiment. Here maximum error and average error are maximum value and average value of error between curve that has mean of candidate control points as Bézier control points and image outline.

Table 1. Result of Bézier curve approximation for seg2 in Fig. 2c

| Algorithm | Iteration number of removing candidate control point | Mean of control point 1 | Variance of control point 1 | Mean of control point 2 | Variance of control point 2 | Maximum error | Average error |
|---|---|---|---|---|---|---|---|
| Method [6] | 0 | (224,187) | (227,153) | (134,38) | (227,153) | 6.11 | 0.79 |
| | 1 | (225,185) | (156,132) | (134,39) | (156,132) | 6.26 | 0.79 |
| | 2 | (224,184) | (151,108) | (135,40) | (151,108) | 6.55 | 0.78 |
| Proposed method | 0 | (218,184) | (82,105) | (128,35) | (129,88) | 5.56 | 0.63 |
| | 1 | (219,184) | (17,98) | (127,35) | (38,79) | 5.86 | 0.63 |
| | 2 | (219,184) | (15,88) | (126,35) | (21,76) | 6.05 | 0.64 |

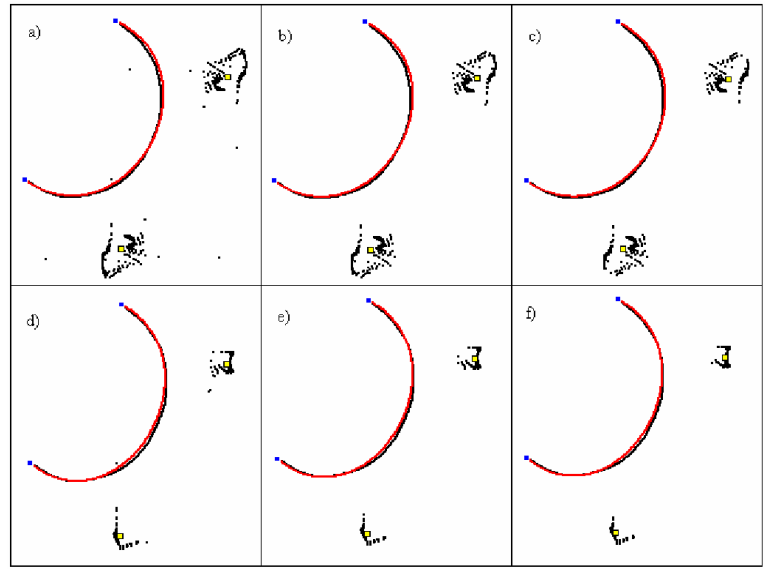

Fig. 3. Control point spread: a) result of [6] by whole control point spread, b) result of [6] by control point spread after iteration 1, c) result of [6] by control point spread after iteration 2, d) result of proposed method by whole control point spread, e) result of proposed method by control point spread after iteration 1 and f) result of proposed method by control point spread after iteration 2.

Table 1 and Fig. 3 show that variance of candidate control point spread, maximum error and average error of proposed method are smaller than previous method. Next, we subdivide seg2 in Fig. 3c. Threshold of control point spread radius is taken as 10 pixels like [6]. Fig. 4 shows subdivided Bézier curve.

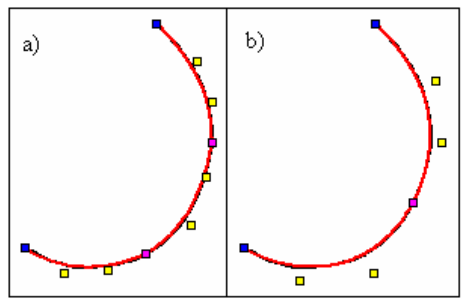

Fig. 4. Subdivided segmentation: a) method [6] and b) proposed method.

In Fig. 4, black curve is image outline, red curve is approximated Bézier curve, blue points are two end points of curve, purple points are end points of subdivided curve and yellow points are Bézier control points of subdivided curves. Maximum errors are respectively 1.86 and 3.52 for method [6] and proposed method and average error are respectively 0.48 and 0.63. As Fig. 4, curve is divided to 3 segments by method [6] but to 2 segments by proposed method for a threshold of control points spread radius. This shows that proposed method guarantee the given accuracy with a small number of subdivision because it has high accuracy.

Next, we do experiment for closed loop curve. Experiment parameters are setting as follows like [6]. $L$ and $R$ is taken as 14, $D$ is taken as 2.6, removing rate of candidate control points for decreasing variance of candidate control point spread is taken as 5%, its iteration number is taken as 2 and threshold of control point spread radius is taken as 10. We applied Bézier approximation method [6] and proposed Bézier approximation method for subdivided segment with subdivision method of [6] to verify strictly the accuracy of proposed method. Fig. 5 shows the result.

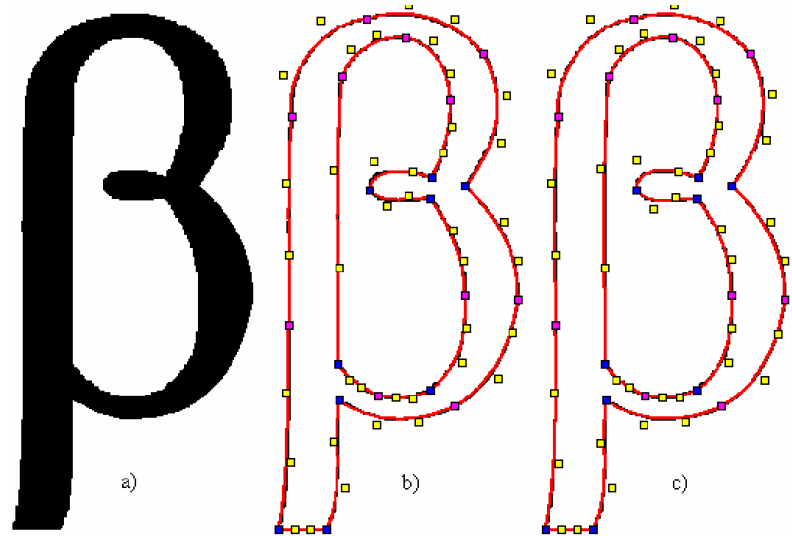

Fig. 5. Comparative experiment: a) original shape, b) result of [6] and c) result by proposed method.

Fig. 5a is original shape of object, Fig. 5b is result of [6] and Fig. 5c is result of proposed method. In Fig. 5b and Fig. 5c, black curve is image outline, red curve is approximated Bézier curve, blue points are two end points of curve, purple points are end points of subdivided curve and yellow points are Bézier control points of subdivided curves. The number of detected corner points is 9. The number of subdivided curves is respectively 20 because we used subdivision method of [6] in two cases. For objectivity of comparative experiment, comparative indicator is set like [6] and the result is showed in Table 2.

Table 2. The result of comparative experiment

| Algorithm | No. of segs. | Compression ratio | Max dev. | Avg. error | Computation time (s) |
|---|---|---|---|---|---|
| Method [6] | 20 | 80.60 | 2.51 | 0.82 | 0.91 |
| Proposed method | 20 | 80.60 | 2.01 | 0.77 | 0.91 |

The curve computation time is iteration time of 10 times including drawing time. From Table 2, we can understand that maximum error and average error of proposed method are smaller than [6]. Next, we applied method [6] and proposed method for shape of Fig. 2. Fig. 6 shows generated Bézier curve and the control points.

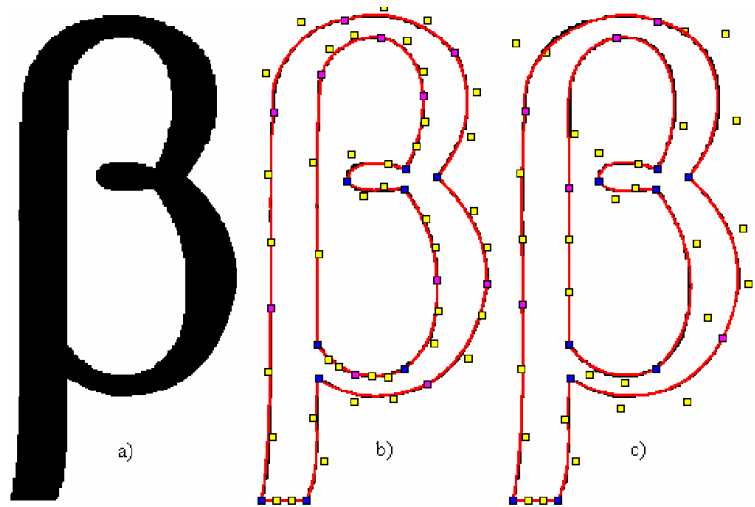

Fig. 6. Comparative experiment: a) original shape, b) result of [6] and c) result by proposed method.

Fig. 6a is original shape of object, Fig. 6b is result of [6] and Fig. 6c is result of proposed method. In Fig. 6b and Fig. 6c, black curve is image outline, red curve is approximated Bézier curve, blue points are two end points of curve, purple points are end points of subdivided curve and yellow points are Bézier control points of subdivided curves. The number of detected corner points is 9. 20 subdivided segments were generated by method [6] and 15 subdivided segments were generated by proposed method after 9 segments was subdivided. Comparative indicator is set like [6] and the result is shown in Table 3.

Table 3. The result of comparative experiment

| Algorithm | No. of segs. | Compression ratio | Max dev. | Avg. error | Computation time (s) |
|---|---|---|---|---|---|
| Method [6] | 20 | 80.60 | 2.51 | 0.82 | 0.91 |
| Proposed method | 15 | 107.47 | 3.52 | 1.22 | 0.72 |

Maximum error and average error of proposed method are larger than [6] in the experiment, because curve is subdivided not to exceed the error limit. The curve computation time is iteration time of 10 times including drawing time as forwards. From Table 3 we can understand that proposed method is better than [6] for data compression rate and computation time because proposed method is very accurate. Comparative experiment results for different shape used previous methods are showed in following figure and table. Experiment condition is like as forwards.

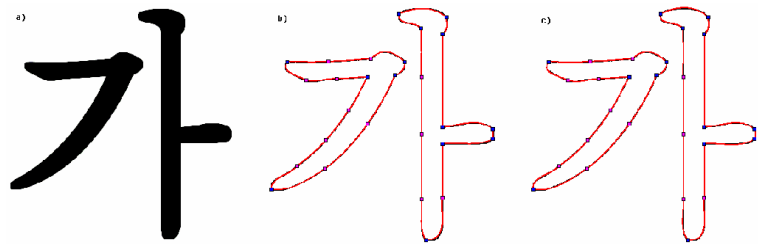

Fig. 7. Comparative experiment: a) original shape, b) result of [6] and c) result by proposed method.

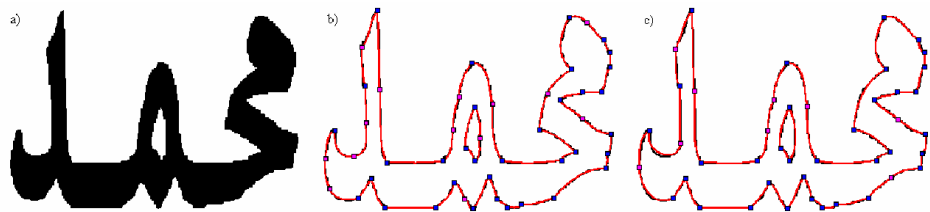

Fig. 8. Comparative experiment: a) original shape, b) result of [6] and c) result by proposed method.

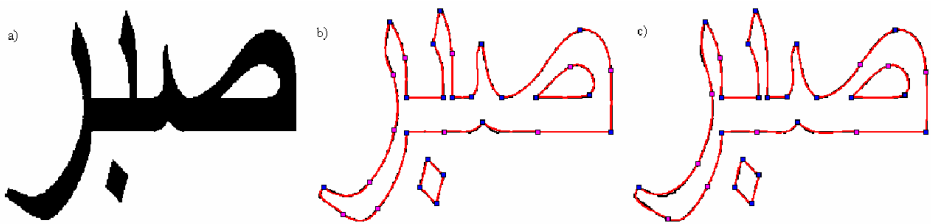

Fig. 9. Comparative experiment: a) original shape, b) result of [6] and c) result by proposed method.

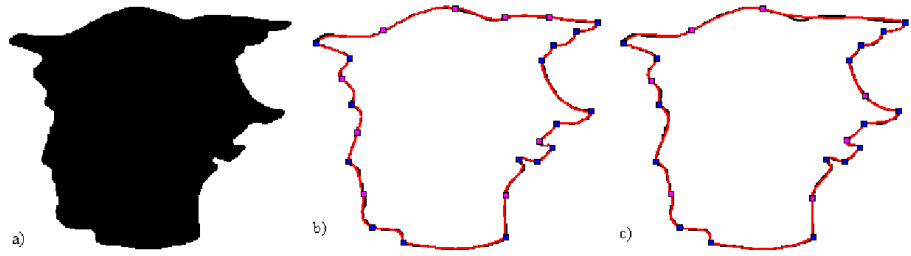

Fig. 10. Comparative experiment: a) original shape, b) result of [6] and c) result by proposed method.

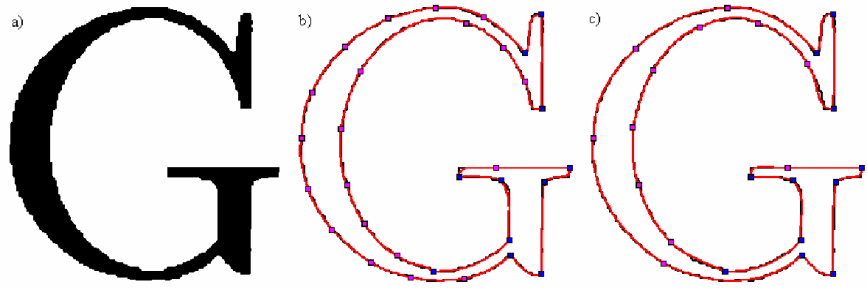

Fig. 11. Comparative experiment: a) original shape, b) result of [6] and c) result by proposed method.

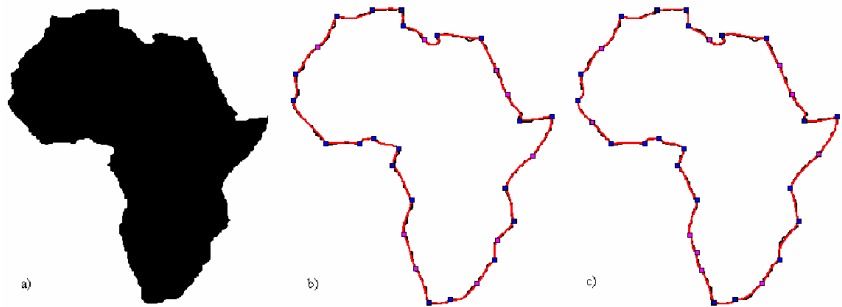

Fig. 12. Comparative experiment: a) original shape, b) result of [6] and c) result by proposed method.

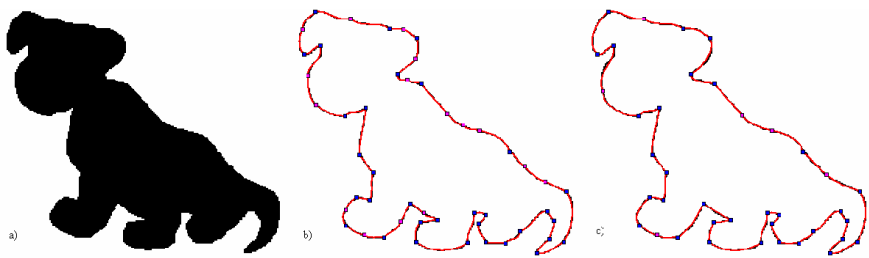

Fig. 13. Comparative experiment: a) original shape, b) result of [6] and c) result by proposed method.

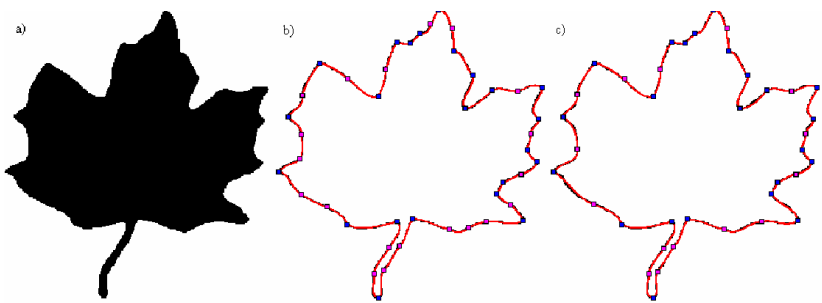

Fig. 14. Comparative experiment: a) original shape, b) result of [6] and c) result by proposed method.

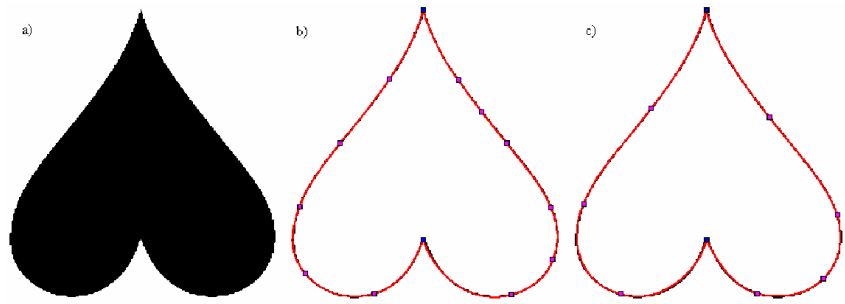

Fig. 15. Comparative experiment: a) original shape, b) result of [6] and c) result by proposed method.

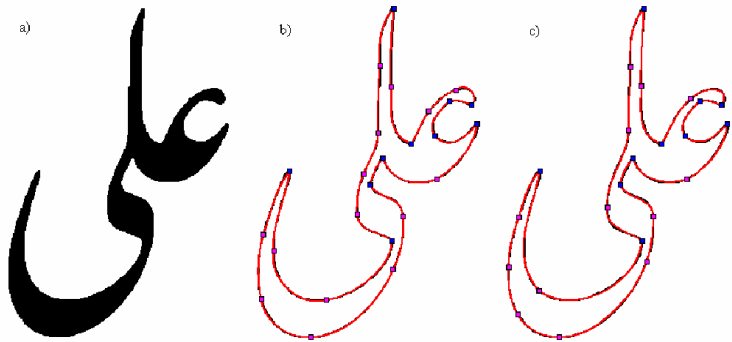

Fig. 16. Comparative experiment: a) original shape, b) result of [6] and c) result by proposed method.

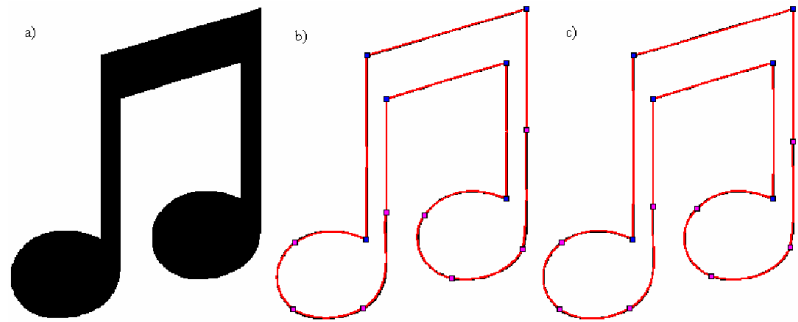

Fig. 17. Comparative experiment: a) original shape, b) result of [6] and c) result by proposed method.

In Fig. 7 ~ Fig. 17, black curves is image outlines, red curves is approximated Bézier curves, blue points are two end points of curves, purple points are end points of subdivided curves. From experiment result, we can understand that data compression rate of proposed method is higher than [6] except Fig. 7, Fig. 12 and Fig. 17. Data compression rate is lower than [6] in the case of Fig. 12 but proposed method is better than [6] for maximum error, average error and computation time. Data compression rate is also equal to [6] in the case of Fig. 7 and Fig. 17 but proposed method is better than [6] for maximum error, average error and computation time. Finally we can understand that proposed method is better than [6] for data compression rate and computation time because proposed method is very accurate.

Table 3. The result of comparative experiment

| Fig. Number | Algorithm | No. of segs. | Compression ratio | Max dev. | Avg. error | Computation time (s) |
|---|---|---|---|---|---|---|
| Fig. 7 | Method [6] | 31 | 45.31 | 3.79 | 0.49 | 1.85 |
| | Proposed method | 31 | 45.31 | 2.43 | 0.47 | 1.84 |
| Fig. 8 | Method [6] | 49 | 40.14 | 3.28 | 0.62 | 0.42 |
| | Proposed method | 49 | 40.14 | 3.28 | 0.62 | 0.42 |
| Fig. 9 | Method [6] | 35 | 82.54 | 2.77 | 0.68 | 0.65 |
| | Proposed method | 34 | 84.97 | 3.31 | 0.76 | 0.63 |
| Fig. 10 | Method [6] | 25 | 36.16 | 2.56 | 0.56 | 0.27 |
| | Proposed method | 23 | 39.30 | 3.45 | 0.69 | 0.25 |
| Fig. 11 | Method [6] | 31 | 51.87 | 3.73 | 0.57 | 1.20 |
| | Proposed method | 23 | 69.91 | 3.48 | 0.61 | 0.98 |
| Fig. 12 | Method [6] | 30 | 34.53 | 3.33 | 0.75 | 0.24 |
| | Proposed method | 31 | 34.42 | 3.26 | 0.68 | 0.22 |
| Fig. 13 | Method [6] | 49 | 35.27 | 2.34 | 0.62 | 0.42 |
| | Proposed method | 39 | 44.31 | 3.80 | 0.72 | 0.34 |
| Fig. 14 | Method [6] | 43 | 40.67 | 2.53 | 0.69 | 0.45 |
| | Proposed method | 38 | 46.03 | 3.41 | 0.74 | 0.45 |
| Fig. 15 | Method [6] | 13 | 72.69 | 1.87 | 0.62 | 0.77 |
| | Proposed method | 9 | 105.00 | 3.05 | 0.72 | 0.65 |
| Fig. 16 | Method [6] | 25 | 60.04 | 2.64 | 0.60 | 0.62 |
| | Proposed method | 23 | 65.26 | 3.75 | 0.69 | 0.78 |
| Fig. 17 | Method [6] | 14 | 110.21 | 1.89 | 0.38 | 0.80 |
| | Proposed method | 14 | 110.21 | 1.71 | 0.37 | 0.76 |

## 6. Conclusion

In this paper, we proposed a technique which determines the mathematic models approximating with cubic Bézier spline for given non-parametric curve such as image outline. The segmentation of given outline uses corner detection and the cubic Bézier spline is obtained from cubic Bézier curve segments. The control points for cubic

Bézier curves approximating segments are determined by using the properties of cubic Bézier curves. The experiment results that the approximation error is much smaller than the previous techniques are shown. The proposed technique can be very useful in various applications that involve problems of cubic Bézier spline approximation for given non-parametric curve such as outline shapes of natural objects.